\documentclass[conference]{IEEEtran}
\IEEEoverridecommandlockouts

\usepackage{graphicx}
\usepackage{textcomp}

\usepackage{setspace}

\usepackage{pifont}

\usepackage{url}
\usepackage{cite}

\usepackage{mathtools}
\usepackage{multirow}
\usepackage{comment}
\usepackage{booktabs}
\usepackage[ruled,vlined]{algorithm2e}
\usepackage{subfigure}
\usepackage{algpseudocode}
\usepackage{listings}

\usepackage{lscape}

\usepackage{tikz,pgfplots}
\usepackage{pgfplotstable}

\usepackage{cite}
\usepackage{amsmath,amssymb,amsfonts}

\usepackage{textcomp}
\usepackage{xcolor}

\def\BibTeX{{\rm B\kern-.05em{\sc i\kern-.025em b}\kern-.08em
    T\kern-.1667em\lower.7ex\hbox{E}\kern-.125emX}}
    
\IEEEoverridecommandlockouts\IEEEpubid{\makebox[\columnwidth]{ 978-1-6654-3540-6/22/\$31.00 ~\copyright~2022 IEEE \hfill} \hspace{\columnsep}\makebox[\columnwidth]{ }}

\begin{document}

\title{Semantic Preserving Adversarial Attack Generation with Autoencoder and Genetic Algorithm}

\author{\IEEEauthorblockN{ Xinyi Wang}
\IEEEauthorblockA{\textit{School of Info. Tech and Elec. Engr.} \\
\textit{University of Queensland}\\
Brisbane, Australia \\
x.wang9@uq.net.au}
\and
\IEEEauthorblockN{Simon Yusuf Enoch}
\IEEEauthorblockA{\textit{Department of Computer Science} \\
\textit{Federal University Kashere}\\
Gombe, Nigeria \\
https://orcid.org/
0000-0002-0970-3621}
\and
\IEEEauthorblockN{ Dan Dongseong Kim}
\IEEEauthorblockA{\textit{School of Info. Tech and Elec. Engr.} \\
\textit{University of Queensland}\\
Brisbane, Australia \\
https://orcid.org/0000-0003-2605-187X}
}

\maketitle

\begin{abstract}
Widely used deep learning models are found to have poor robustness. Little noises can fool state-of-the-art models into making incorrect predictions. While there is a great deal of high-performance attack generation methods, most of them directly add perturbations to original data and measure them using L\_p norms; this can break the major structure of data, thus, creating invalid attacks. In this paper, we propose a black-box attack, which, instead of modifying original data, modifies latent features of data extracted by an autoencoder; then, we measure noises in semantic space to protect the semantics of data. We trained autoencoders on MNIST and CIFAR-10 datasets and found optimal adversarial perturbations using a genetic algorithm. Our approach achieved a 100\% attack success rate on the first 100 data of MNIST and CIFAR-10 datasets with less perturbation than FGSM. 
\end{abstract}


 \footnotetext[1]{\textbf{Cite this article as:}\\ Wang, X., Enoch, S. Y., and Kim, D. S. (2022). Semantic Preserving Adversarial Attack Generation with Autoencoder and Genetic Algorithm. \textit{In 2022 IEEE Global Communications Conference (GLOBECOM).}IEEE, 2022}
	
\section{Introduction}

The rapid growth of artificial intelligence (AI) techniques has popularized deep learning (DL) for handling machine learning tasks such as malware classification, image classification, spam detection, \textit{etc}. However, the DL models are known to be vulnerable to attacks \cite{hao2022adversarial}. Samples with carefully designed noises, namely adversarial attacks, have been found to be able to easily fool state-of-the-art DL models. This vulnerability to adversarial attacks is found to be a general problem among many Machine Learning (ML) and DL architectures. As an example, one can use 3D printed eyeglasses to fool the facial recognition system into regarding him or her as another person \cite{sharif2016accessorize}; therefore, achieving an attack on the facial recognition payment system. Another example is self-driving cars; slightly modified traffic signs can mislead them into making incorrect decisions \cite{morgulis2019fooling} therefore resulting in traffic accidents.

Given these challenges, techniques for attack generation and defense have been researched and proposed in several studies. However, most of the attack methods use L\_p norms for measurement of perturbations and they focused on generating attacks with small L\_p distances, while less attention is paid to the semantics of samples. Providing that noises with small L\_p norms can still produce perceptually different adversarial examples \cite{sharif2018suitability}, this work focuses on generating attacks with small perturbations while preserving the semantics of original samples.

In this paper, we proposed a new approach for adversarial attack generation which uses an autoencoder to preserve the semantics of data and finds optimal perturbations with a genetic algorithm. In particular, we proposed a new black-box attack approach to achieve the best attack success rate on the data of MNIST (Modified National Institute of Standards and Technology dataset) and CIFAR-10 (Canadian Institute For Advanced Research) datasets with relatively smaller perturbation sizes than FGSM \cite{goodfellow2014explaining}. Here, the MNIST dataset is a dataset of 60,000 small square 28×28 pixel grayscale images of handwritten single digits between 0 and 9, while the CIFAR-10 dataset consists of 60,000 32x32 color images in 10 classes, with 6,000 images per class. There are 50,000 training images and 10,000 test images.
Both datasets are among the most widely used datasets for machine learning research. The main contributions of this paper are as follows:

\begin{itemize}
    \item We proposed a new approach for adversarial attack generation using an autoencoder to preserve the semantics of data.
    \item We compute the optimal perturbations for the attack using a genetic algorithm.
    \item We evaluate the performance of untargeted and targeted attacks and then compared them with several basic and state-of-the-art attacks.
  
\end{itemize}

The rest of the paper is structured as follows. Section \ref{sec:relatedwork} provides the related work and the background to the study. 
Our proposed approach is presented in Section \ref{sec:proposed_approach}. 
In Section \ref{sec:experiments}, we provide the description of the experimental setup and analysis of the obtained results. The conclusion, discussion, and future work are provided in Section \ref{sec:conclude}.

\section{Related work}
\label{sec:relatedwork}

In this section, we review relevant papers based on methods for generating attacks for text classifiers and image classifiers. 

\textbf{Attacks to image classifiers:} Goodfellow \textit{et al.} \cite{goodfellow2014explaining} proposed the Fast Gradient Sign Method (FGSM). This method achieves attack by moving a seed towards the direction in which the loss of a target model increases. It supports untargeted attacks and can be modified to generate targeted attacks by calculating noises to minimize loss with regard to target classes. Kurakin \textit{et al.} \cite{kurakin2016adversarial} proposed a basic iterative method based on the same hypothesis of FGSM (the linearity hypothesis). Different from FGSM, BIM is an iterative method that adds small noises to seeds iteratively. At each step, noises are calculated with the same equation as FGSM.

Szegedy \textit{et al.} \cite{szegedy2013intriguing} defined adversarial attack generation as an optimization problem and solved it with the Limited-Memory BFGS algorithm. The authors applied this attack on different neural networks and different datasets; they successfully generated attacks with little noise in all experiments. Similarly, Carlini \textit{et al.} \cite{carlini2017towards} also defined generation of attacks as an optimization problem. Different from L-BFGS, they explored seven different objective functions to overcome the non-linearity of the objective function in the L-BFGS attack. 

\textbf{Attacks to text classifiers:}
Generating attacks on text classifiers is different from generating attacks on image classifiers. Adding noises to texts as to images will lead to invalid words or sentences. 
Li \textit{et al}.\cite{li2018textbugger} proposed TextBugger which combines character level perturbation with word-level perturbation to generate attacks. For word-level perturbation, they first calculate words’ importance using the Jacobian matrix, then, closest neighbors of important words are found to replace them using a pre-trained model which embeds word semantics. In character-level perturbation, important words are found in the same way as in word-level attacks. Alzantot \textit{et al.} \cite{alzantot2018generating} proposed a word-level attack using genetic algorithm. The key idea of their attack is to replace as few as possible words with semantically similar words. They use GloVe embedding space to measure the semantic distance of words to find some candidates; then, Google's 1 billion words language model is used to check if these candidates are suitable in original sentences. Optimal replacements are found with a genetic algorithm by checking how wrong the target model is; that is, the confidence with which the target model misclassifies an attack. 

\vspace*{-0.1cm}

\section{The Proposed Approach}
\label{sec:proposed_approach}

We proposed a black-box approach for adversarial attack generation trying to provide a semantic preservation functionality. We use latent space learned by an autoencoder to measure the semantic distance between attacks and their seeds and use GA to find optimal noises. The major characteristics of our method are; Black-box, Support for untargeted and targeted attacks, No direct modification of original data, and considering semantic distance during attack generation

Similar to L-BFGS, we define attack generation as an optimization problem but with an additional term to minimize: 

\begin{align*} 
&\textnormal{Minimize: } D_1(x,x\prime) + D_2(o,o\prime)\\
&\textnormal{subject to: }f\left(x\prime\right) !=y \\
& \textnormal{and}~~ x\prime\in\ [l,r]n \\
 & \textnormal{where}:o=encode\left(x\right),~o\prime = o + \delta,~x\prime = decode(o + \delta),\\
~&\textnormal{and}~f(x) = y
\end{align*}

\begin{figure*}[!h]
	\centering
	\includegraphics[width=0.7 \linewidth]{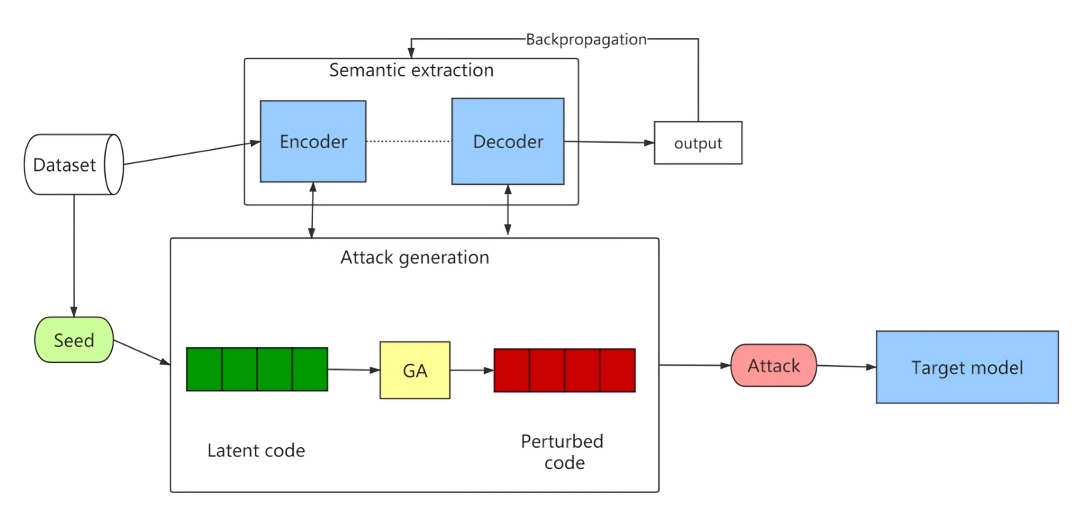}
	\caption{The proposed approach}
	\label{fig:proposed_approach}
\end{figure*}

where x is a seed, y is the correct class of x, f is the classifier under attack, l and r define the valid range of data, n is the dimension of data, $x\prime$ is an attack, $D_1$ and $D_2$ are distance functions, o and $o\prime$ are the latent code of x and $x\prime$,$\delta$ is the noise in latent space. That is, for a seed x, we want to find the optimal attack $x\prime$ which is close to x in both original space and semantic space. We generate such $x\prime$ by adding noises on the latent code of x and decode the perturbed latent code with the decoder. We use GA to solve this optimization problem. The overall procedure of our approach is shown in Figure \ref{fig:proposed_approach}.

Our approach supports both untargeted and targeted attacks. Furthermore, with the help of an autoencoder, our approach is possible to be extended to work on other forms of datasets apart from image datasets.

For image data, we use L\_2 norm to measure both pixel distance and semantic distance between attacks and their seeds. This is because noises with small L\_2 norm are less likely to distort semantics of images according to the experiment results from \cite{sharif2018suitability}. Therefore, assuming pixels are normalized to [0, 1], the problem becomes:
\begin{align*}
&\textnormal{Minimize: } ||\mu||_2 + ||\delta||_2\\
&\textnormal{subject to: }f\left(x\prime \right) !=y \\
& \textnormal{and}~x\prime\in [0,1]^n, 
& \textnormal{where:}  \mu = x - x\prime
\end{align*}  
For other forms of data, problem-specific distance can be used to replace L\_2 distance.

\subsection{Feature extraction}

The intuition behind our method is to protect major features of seeds to preserve their semantics and core structures. Because the latent space of an autoencoder is believed to be a semantic space, we use this space to measure the semantic distance to achieve our functionality. Training an autoencoder on data, we get:
(1) A latent space for semantic distance measurement, (2) A low dimensional space for GA to search on, and 	(3) A decoder to generate attacks. After training, we obtain two functions: 
\begin{align*}
Encode(x) = o, ~~\textnormal{and} 
& ~~Decode(o) = x\prime
\end{align*}
The decoder is then used as a generative model; new samples are generated using the decoder by modifying latent code o.

\subsection{Attack generation}

\begin{figure*}[!h]
	\centering
	\includegraphics[width=0.9\linewidth]{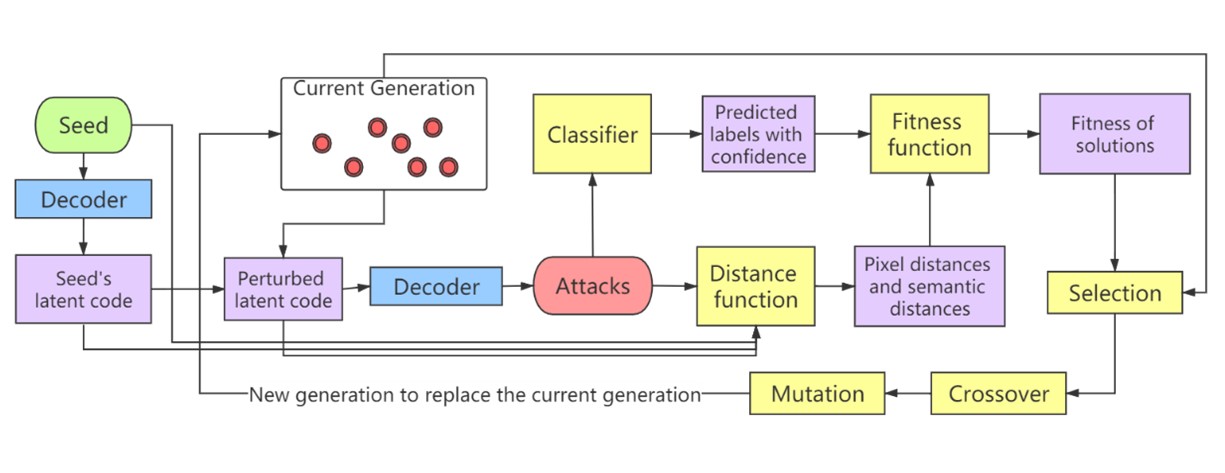}
	 \vspace*{-0.5cm}
	\caption{The overall procedure of GA}
	\label{fig:GA_steps}
\end{figure*}

After we have extracted the latent space, attack generation can be done by a genetic algorithm. Given a seed x, we can easily get its latent code using the decoder. Then, we use chromosomes to represent noises and search for optimal noises in the latent space. The main procedure of attack generation is illustrated in Figure \ref{fig:GA_steps}.

\subsubsection{Chromosome}

\begin{figure}[!htb]
	\centering
	\includegraphics[width=1.0\linewidth]{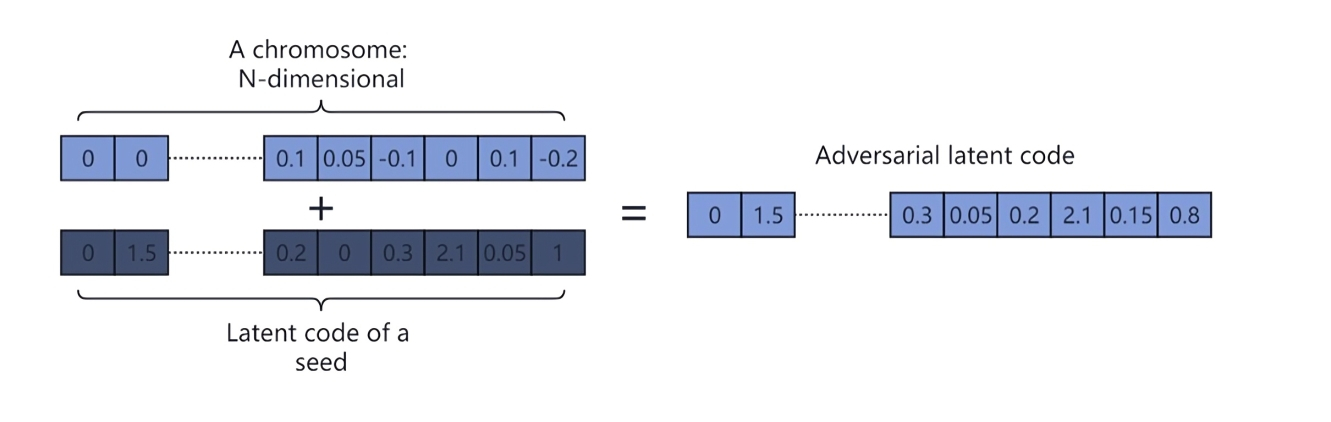}
	\vspace*{-0.8cm}
	\caption{Process of constructing an adversarial code with a chromosome}
	\label{fig:chromosome}
\end{figure}

We encode noises on latent space into chromosomes; therefore, a chromosome is of the same dimension as the latent space of the autoencoder. Each gene in a chromosome is a float value which is the noise to the seed’s latent code at the same location. Given a seed x and a chromosome c, a new sample can be generated with the decoder (Figure \ref{fig:chromosome}) :~~~ $x^\prime=decode(encode(x) + c)$

\subsubsection{Selection}

We use elitism selection in GA for offspring production. In each generation, the best k individuals are protected and added to the next generation without modification. Other individuals of the next generation are produced by Roulette wheel selection. Note that the elitists are also considered when choosing parents so that their good genes can be shared by others.

\subsubsection{Crossover}

We use one-point crossover in this work to keep the implementation and parameter tuning simple. When crossover happens, a random position is chosen and then used to split and recombine parents. 

\subsubsection{Mutation}

We apply mutation operation on new offspring after crossover. Once a gene is decided to be mutated, it will get modified with a pre-defined step size. Because genes are of float type, this step size is a small float value and its magnitude is tuned individually for different datasets.

\subsubsection{Fitness function}

Fitness describes how good a perturbation is. We evaluate adversarial perturbations from three aspects: (1) perturbation size, (2) semantic distance, and (3) attack performance. 

Perturbation size is the distance between an attack and its seed; given a sample x and chromosome c:
$PS(x, c)=||decode(h+c)-decode(h)||_2$
where $h = encode(x)$ is the latent code of sample $x$. Instead of the original x, the reconstructed sample ($decode(h)$) is used in the equation; this is because the distance between $x$ and $decode(h)$ may make the objective of GA become decreasing the loss of the autoencoder instead of finding adversarial attacks.

Semantic distance is the L\_2 norm of perturbations in the latent space; given a chromosome c:
$SD(c)=||c||_2$ 
Evaluation of attack performance uses confidence of the target model which is inspired by \cite{chen2019poba}. As for the untargeted attack, given a seed x and a chromosome c, then it can be computed by Equation \eqref{eq_attack_performance_untargeted}.

\begin{equation}\label{eq_attack_performance_untargeted}
\begin{gathered} 
AP(y,p)=\left\{
                \begin{array}{ll}
                \max(p) - second\_max(p), & y!=y\prime \\
                  -(max(p) - second\_max(p)), &  y==y\prime\\
                \end{array}
              \right.
\end{gathered}
\end{equation}

where y is the true label of sample x and $y\prime$ is the predicted label of $x\prime$ which is generated by c: $x\prime$ = $decode(encode(x) + c)$; p is the output of target model where $p[yi]$ is the probability of $x$ belonging to $yi$. When a sample is correctly classified by the target model, we want to decrease the model’s confidence in this prediction; when a sample is misclassified, we want to increase the model’s confidence in the wrong prediction, therefore, making the attack more aggressive.

For a targeted attack, the probability of x belonging to class $y\prime\prime$ is used to direct the search as given by Equation \eqref{eq_attack_performance_targeted}.

\begin{equation}\label{eq_attack_performance_targeted}
\begin{gathered} 
AP(y\prime\prime,p)=\left\{
                \begin{array}{ll}
                \max(p) - second\_max(p), & y\prime=y\prime\prime \\
                  -(max(p) - p[y\prime\prime]), &  y\prime!==y\prime\prime\\
                \end{array}
              \right.
\end{gathered}
\end{equation}

where $y\prime\prime$ is the targeted class label, $p[y\prime\prime]$ is the probability of x belonging to class $y\prime\prime$, and $y\prime$ is the predicted label.

Given above definition, fitness function for a chromosome c is defined by Equation \eqref{eq_fitness_function}.

\begin{equation}\label{eq_fitness_function}
\begin{gathered} 
Fitness=\left\{
                \begin{array}{ll}
                AP-PS\times \alpha - SD \times \beta, & (\text{misclassified}) \\
                  AP, &  (\text{correctly classify})\\
                \end{array}
              \right.
\end{gathered}
\end{equation}

The hyper-parameter $\alpha$ and $\beta$ are used to control the importance of perturbation size and semantic distance. With small $\alpha$ and $\beta$, the searching process will focus more on attack performance, while with large $\alpha$ and $\beta$, attacks with large PS and SD will be eliminated and the search focus more on perturbation size; therefore, using these two parameters, we can have a trade-off between attack performance and perturbation size. Perturbation size and semantic distance are not considered if a generated sample is correctly classified; this allows GA to keep the noises that are most likely to create adversarial attacks when there is no attack found, therefore reaching the space in which attacks lay faster.

\section{Experiment, Results, and analysis}
\label{sec:experiments}

We choose to use two well-known datasets MNIST digits and CIFAR-10 colored images. 

\subsection{Experiments on MNIST}
\subsubsection{Autoencoder }

\begin{figure}[!htb]
	\centering
	\includegraphics[width=0.7\linewidth]{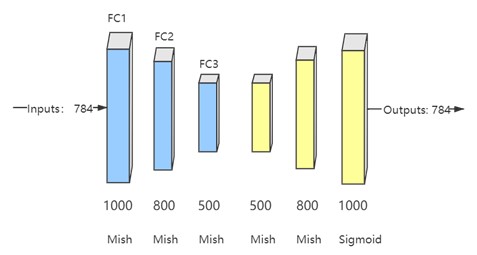}
	\caption{Structure of initial autoencoder; the number of nodes of each layer is later reduced to 512, 264, 128, 128, 264, 512}
	\label{fig:initial_autoencoder}
\end{figure}

We trained a simple sparse autoencoder on the training set of MNIST. Firstly, we built an autoencoder with only fully connected layers and found it performs quite well on MNIST. The structure of our initial autoencoder is shown in Figure \ref{fig:initial_autoencoder}.

\begin{figure}[!h]
\vspace*{-0.1cm}
	\centering
	\subfigure[Training loss of the autoencoder]{
		\label{fig:autoencoder_training_a}
		\includegraphics[width=0.4\textwidth]{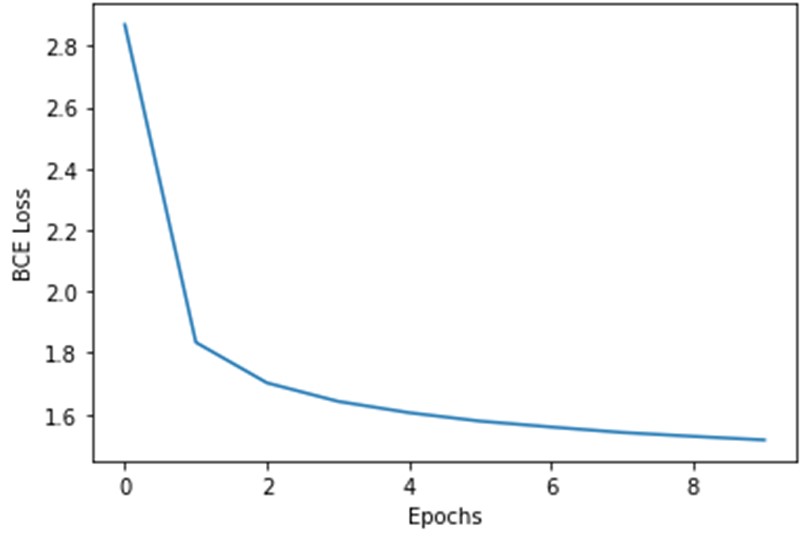}
	}
	\subfigure[Extracted latend code of a sample]{
		\label{fig:autoencoder_training_b}
		\includegraphics[width=0.45\textwidth]{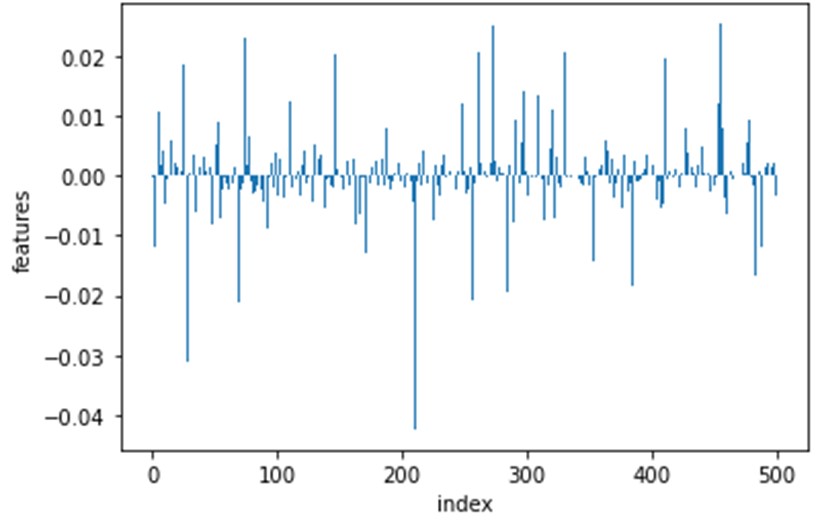} 
	}
	\caption{Results showing the training loss of the autoencoder and the extracted features of a random sample from training set}
	\label{fig:autoencoder_training}
\end{figure}

We use binary cross-entropy loss to calculate reconstruction error and Adam optimizer to train our model because Adam can adjust the learning rate automatically. With a learning rate of 0.001, we trained the autoencoder for 10 epochs in Figure \ref{fig:autoencoder_training}. 

\begin{figure}[h]
	\centering
	\subfigure[ Original from training set]{
		\label{fig:autoencoder_training_4a}
		\includegraphics[width=0.152\textwidth]{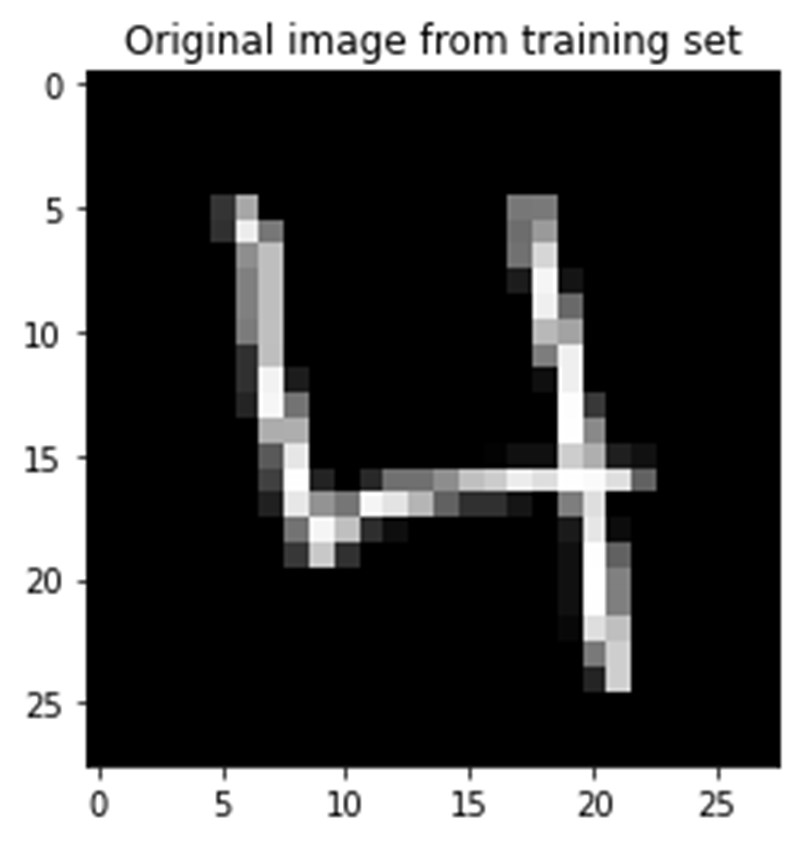}
	}
	\subfigure[ The reconstructed image]{
		\label{fig:autoencoder_training_4b}
		\includegraphics[width=0.152\textwidth]{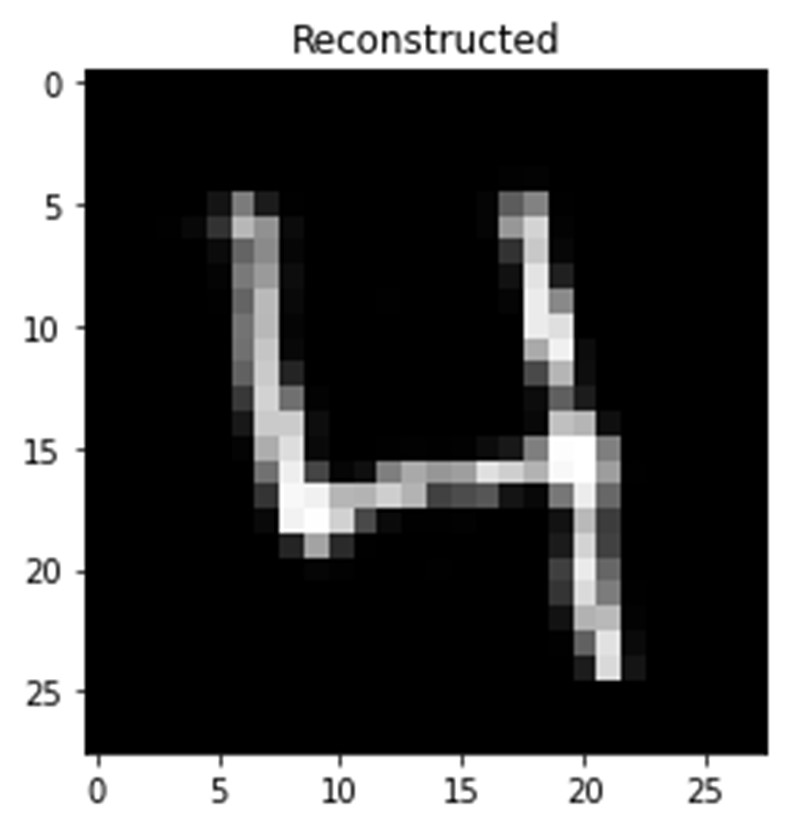} 
	}
	\caption{a random seed from training set and its reconstructed image - 10 epochs}
	\label{fig:autoencoder_training_4}
\end{figure}

\begin{figure}[h]
	\centering
	\subfigure[Training loss of final autoencoder over epochs]{
		\label{fig:final_autoencoder_a}
		\includegraphics[width=0.4\textwidth]{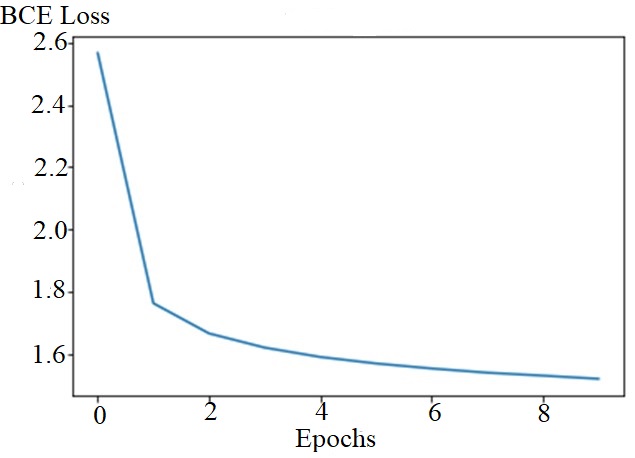}
	}
	\subfigure[Reconstructed images using random seeds from the test set]{
		\label{fig:final_autoencoder_b}
		\includegraphics[width=0.4\textwidth]{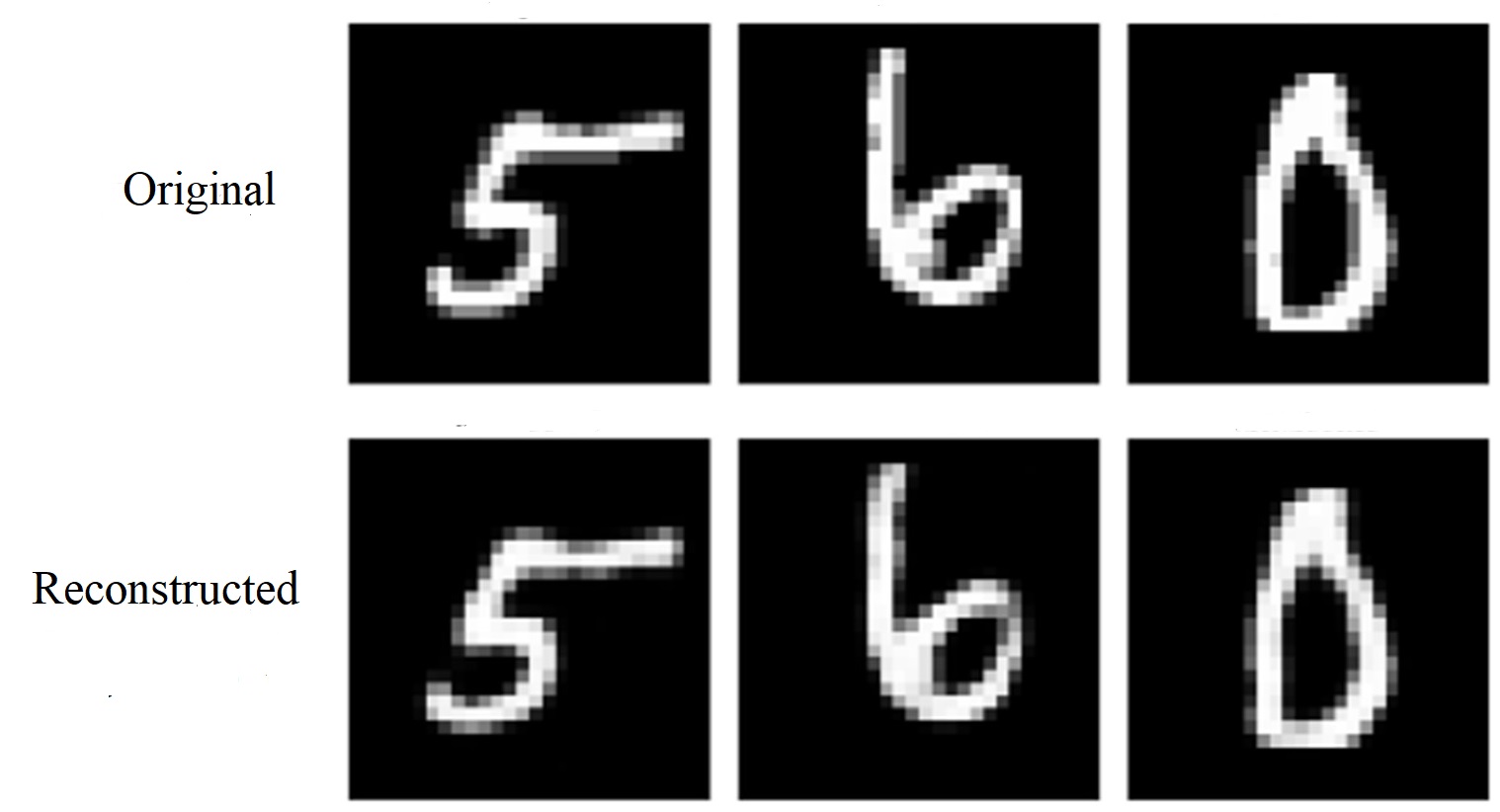} 
	}
	\caption{final autoencoder over epochs}
	\label{fig:final_autoencoder}
\end{figure}

\begin{figure}[!htb]
	\centering
	\includegraphics[width=0.8\linewidth]{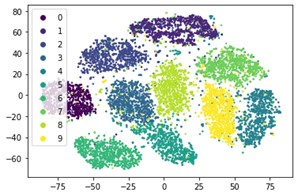}
	\caption{Distribution of testing data in latent space of the final autoencoder (visualized with T-SNE), x and y axis are embeddings of latent codes calculated by T-SNE}
	\label{fig:Distribution_final_autoencoder}
\end{figure}

Then, we tried to reduce the number of nodes in each layer because the dimension of latent space will affect GA’s time efficiency. After exploring different settings, we finally reduced the number of nodes in the latent layer to 128 which have similar reconstruction performance and are small enough so that GA can run fast on it. Because a decrease in the number of nodes in each layer will cause the reconstruction error to increase but the further decrease has little improvement in GA’s time efficiency (from 128 to 64: around 2 seconds on average for 100 generations), we stopped reducing the number of nodes in latent space. The structure of our final encoder is Layer 1: 512 nodes, Layer 2: 264 nodes, and Layer 3: 128 nodes; the decoder has a symmetrical structure. Training loss, examples of reconstructed images, and distribution of latent space is illustrated in Figure \ref{fig:final_autoencoder}, and Figure \ref{fig:Distribution_final_autoencoder}.

We use the Mish activation function in the latent layer because it has been shown to perform better than other activation functions on many datasets \cite{misra2019mish} as it provides a smooth optimization surface. Another feature of the Mish function is that it allows negative values while ReLU does not. Because mutation in GA can result in many negative noises, given most of the latent features are close to 0 in a sparse autoencoder, a lot of noises will be unreasonable; therefore, the Mish function works better in this scenario.

\subsubsection{Model under attack}
\begin{figure}[!h]
	\centering
	\includegraphics[width=1.0\linewidth]{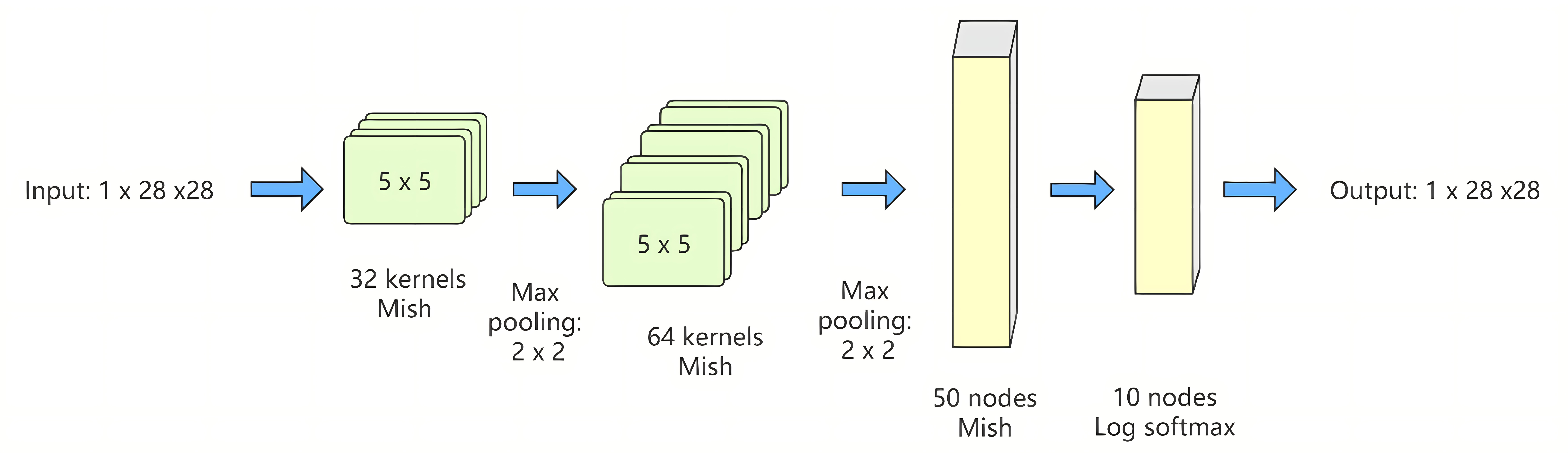}
	\caption{Structure of the target model}
	\label{fig:structure_target_model}
\end{figure}

We trained a convolutional network on the training set of MNIST and used it as a target model for attack generation. The structure of this model is illustrated in Figure \ref{fig:structure_target_model}. After training for 25 epochs with a learning rate of 0.001, the target model achieved 99.05\% accuracy on the test set.

\subsubsection{Attack generation with GA}
With the trained autoencoder, we can start generating attacks with GA. Taking a random image from the training set, we pass the latent code of this seed to GA and modify it with chromosomes. After 100 generations, we successfully got an adversarial attack.
Because there are many hyperparameters in GA, we fixed some of them with the same values heuristically and tune one or two parameters at each time. We split the parameter into the following groups:
\begin{itemize}
	\item Crossover rate and mutation rate; initial value: 0.2 and 0.1
	\item The number of population in each generation and the number elitist to keep; initial value: 50 and 5 respectively
	\item Range of initial noises and step size used in mutation; initial value: [0, 0.04] and 0.02
	\item $\alpha$ in fitness function which controls the importance of perturbation size; initial value: 0.2
	\item $\beta$ in fitness function which controls the importance of semantic distance; initial value: 1
	\item Number of generations before termination; initial value: 100
\end{itemize}
The initial values were decided by generating several attacks and manually checking the generated images as well as the magnitude of perturbations. 

Lastly, using the previously tuned parameters, we generated attacks with different numbers of generations. Average perturbation size goes down when the number of generations increases while average fitness grows up when the number of generations increases. However, fitness increases slowly when the number of generations is larger than 150 while the running time increases quickly. Therefore, we think 150 generations is a good choice. 

\begin{figure}[!h]
	\centering
	\includegraphics[width=0.9 \linewidth]{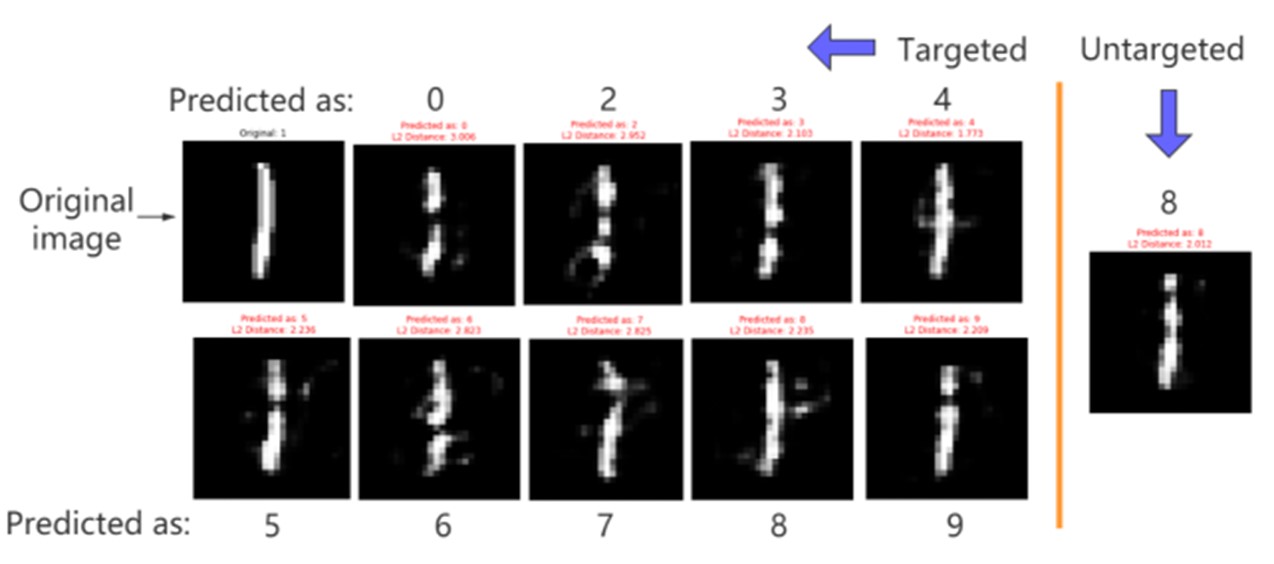}
	\caption{Comparison of untargeted and targeted attack with a random seed from the training set; ($\alpha$ = 0.4)}
	\label{fig:untargeted_vs_targeted_attack}
\end{figure}

After an untargeted attack, we explored targeted attacks with the same hyperparameters. However, we found that targeted attacks exhibit more distortion (Figure \ref{fig:untargeted_vs_targeted_attack}) in comparison to untargeted attacks. This is because GA works on a non-convex fitness function and different initialization can lead to different results; while an untargeted attack finds the best attack starting from its initial population, a targeted attack has to discard many good attacks and is more affected by initialization.

\subsubsection{Comparison with other attacks}
To compare our proposed work with other attack methods, we generated attacks with foolbox \cite{rauber2020foolbox} using the first 100 data from MINST’s training set as seeds. Mean perturbation per pixel is measured using L\_2 norm and the average running time for a single attack is counted in seconds. 

We found that our approach has a good attack success rate as state-of-the-art attacks, but our attack has around 3 times more perturbations than the Boundary attack and C\&W attack. 

Attacks generated by FGSM, BIM, and PGD have more noise than those generated by other attacks. C\&W is one of the most powerful attacks which achieved a 100\% attack success rate with the lowest average perturbation size. Boundary attack also has good performance; being a black-box attack, it achieved the same success rate as C\&W with close perturbation size and even better time efficiency.

Using the first 6 images from MNIST’s training set, we generated adversarial attacks with different attack methods. We noticed that compared to other attacks, those generated by our method have features from other classes, or have their own features weakened; some attacks look like interpolations between different classes. 

\begin{figure}[!h]
	\centering
	\includegraphics[width=0.9\linewidth]{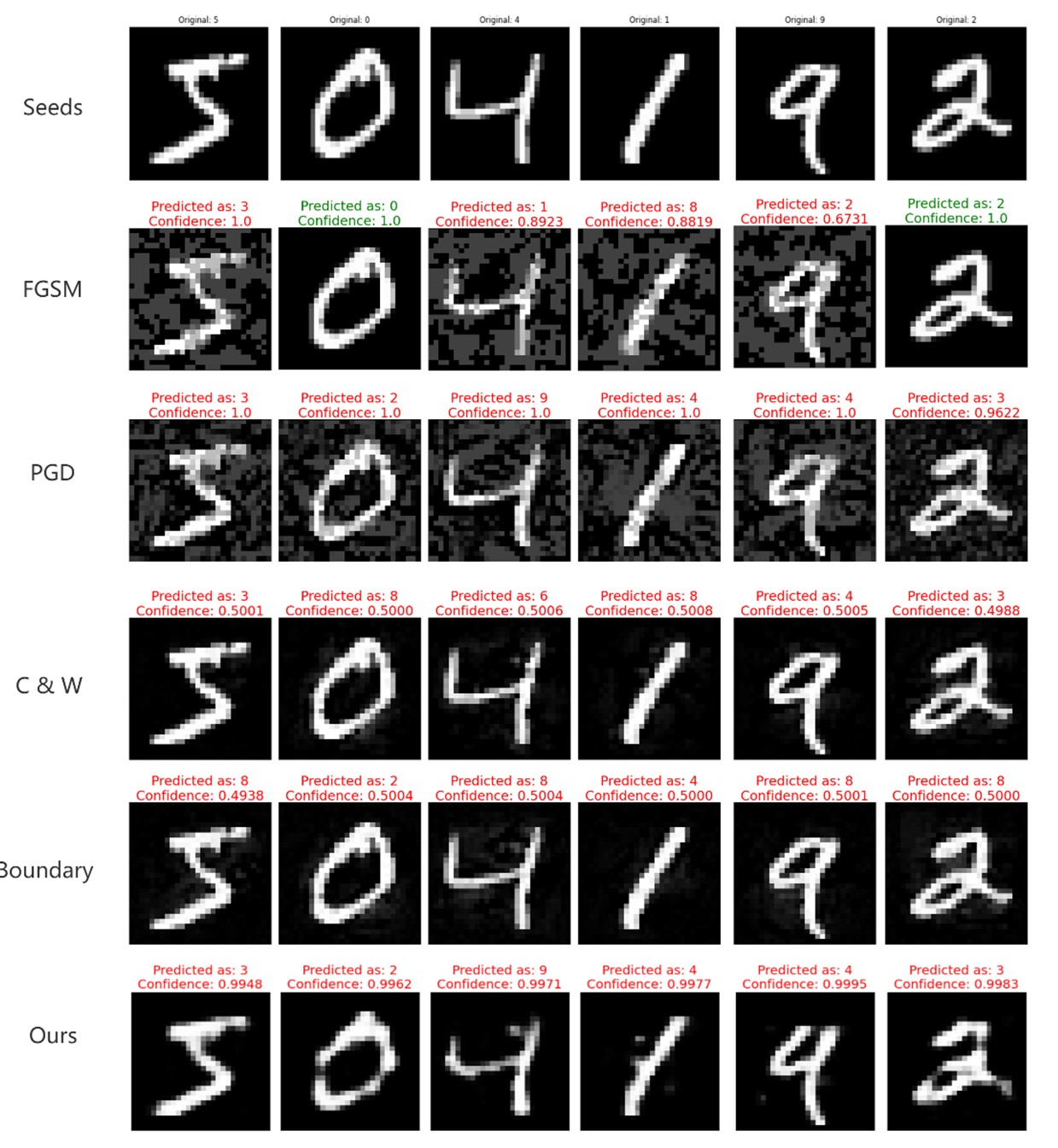}
	\caption{Attacks generated by different methods. The first 6 data from the training set of MNIST are used. The second last attack generated by our approach is overly distorted}
	\label{fig:attack_diffMethods}
\end{figure}

We found an attack that is semantically different from its seed. The attack generated for number ‘9’ in Figure \ref{fig:attack_diffMethods} has its upper part erased; although it is misclassified by the target model with a small perturbation size, it looks fairly close to number ‘4’ and should be considered an invalid attack. This indicates that we might have to further explore the effectiveness of $\beta$ to better understand our method and prevent this kind of attack from being produced. However, this also indicates that our approach is truly working on the semantic space and can potentially provide a semantic preservation functionality. More experiments are needed to improve and validate this functionality.

\subsection{Experiments on CIFAR-10}
\subsubsection{Autoencoder}
Because the CIFAR-10 dataset contains colored images, the dimension of data becomes $3 \times 32\times 32 = 3072$ which is about 4 times larger than that of MNIST. Therefore, instead of a fully connected autoencoder, we used a convolutional autoencoder from \cite{Jie} with some modifications on activation functions. 
We trained the autoencoder with a training set for 30 epochs and observed good reconstruction performance. 

\subsubsection{Model under attack}

We found that simple convolutional networks have poor performance on the CIFAR-10 dataset (around 65\% accuracy on the testing set), therefore, to better understand the performance of our proposed method, we use a pre-trained VGG-11 model from \cite{Phan} which has testing accuracy of 92.39\%.

\subsubsection{Attack generation}

Because the dimension of data and semantic space have changed, hyperparameters tuned for the MNIST dataset do not work well now. Initialization range and step size are affected by the magnitude of latent features and the previously used values cannot find any adversarial attack this time; $\alpha$ value also has to be returned because perturbations have different magnitudes on colored images in relation to grayscale images. 

The crossover rate and mutation rate are kept the same as before because we have the same selection, crossover, and mutation scheme. The effectiveness of population size and the number of elitists to keep per generation is clear according to the previous experiments; increasing them together can improve average fitness but also increase the running time. Therefore, they are left unmodified to keep a balance between the quality of solutions and time efficiency. The number of generations is decreased to 100 to make the later experiments easier; it is set back to 150 when evaluating the overall performance of the algorithm.

Observing the magnitude of latent features, we set the value of $\beta$ to 0.01 heuristically. Also, because the dimension of data is about 4 times higher than before (from 784 to 3072), we set the value of $\alpha$ to 0.1. Keeping other parameters unmodified, we generated 10 attacks with random seeds for different initialization ranges and step sizes. Although 10 attacks are not enough to show a stable average performance and the plots are not smooth, the overall tendency consents to the previous experiments.

\begin{figure}[!h]

	\centering
	\includegraphics[width=0.9\linewidth]{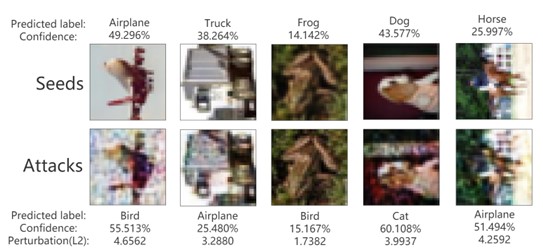}
	\caption{Attacks generated with max initial noise of 1.5 and step size of 0.3 using random seeds from the training set}
	\label{fig:attack_gen_CIFAR}

\end{figure}

We give more importance to attack success rate and leave $\alpha$ to control perturbation size. We choose 1.5 as the max magnitude of initial noises and 0.3 as the step size (Figure \ref{fig:attack_gen_CIFAR}).

Finally, we explored the effectiveness of $\beta$. We found that generated attacks are overly distorted with a $\beta$ of 0, and the attacks are similar with non-zero $\beta$ values. This agrees with the experiment on MNIST. However, to validate the semantic preservation functionality, further experiments are needed. Using the same parameters as MNIST, we generated some targeted attacks. However, we found many of them are overly distorted or failed and only those with their original classes close to targeted classes (cat and dog, automobile and truck, etc) succeeded. We explain this as the use of autoencoder and its latent space; because the map from latent space to image space is not continuous, noises cannot be freely added onto seeds but have to follow the patterns learned by the decoder. This property somehow narrowed the searching space to a small range and led to poor performance on targeted attacks. The over distortion indicates that our choice of $\alpha$ may be too small; it works on the untargeted attack because they are easy to find, but it cannot guarantee that all perturbations are small. Therefore, a targeted attack on CIFAR-10 left as a future job.

\subsubsection{Comparison with other attacks}

Foolbox \cite{rauber2020foolbox} is again used to compare our untargeted attack with other attacks on CIFAR-10. We generated 100 attacks with different methods and collected their performance. The first 100 images from the training set are used as seeds and the result is recorded.

We noticed that our approach has poor time efficiency; the running time is 3 times longer than C\&W and more than 10 times longer than the Boundary attack. 


\begin{figure}[!h]
	\centering
	\includegraphics[width=0.8\linewidth]{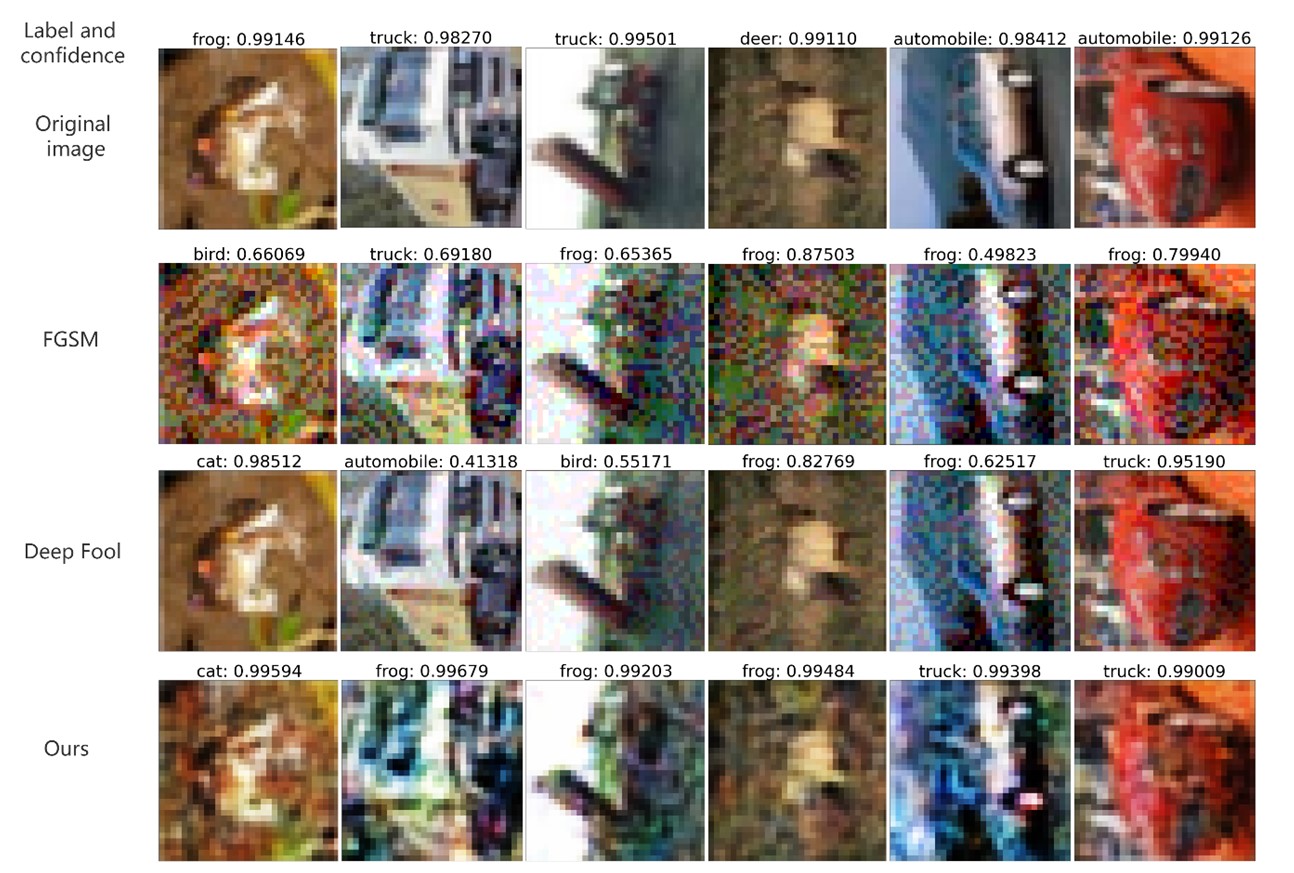}
	\caption{Attacks generated by different algorithms compared with our approach; first 6 images from the training set are used as seeds}
	\label{fig:attack_gen_diffAlgorithm}
\end{figure}

\section{Conclusion}
\label{sec:conclude}

We developed a new black-box attack approach that achieved a 100\% attack success rate on the first 100 data of MNIST and CIFAR-10 datasets with relatively smaller perturbation size than FGSM. While the average perturbation size of our attack is larger than state-of-the-art attacks such as Boundary attacks and C\&W, our attack is more aggressive and most of the attacks are misclassified by the target model with a probability higher than 99\% (Figure \ref{fig:attack_diffMethods}, Figure \ref{fig:attack_gen_diffAlgorithm}). In addition, attacks generated by our method have better diversity; using a single seed, different running of GA can find different attacks with different class labels. This can be helpful to adversarial training. Our approach does not require gradients and is more practical than regular white-box attacks. Although our approach has poor time efficiency compared with other attacks, it can be improved by moving GA to GPU. We have evaluated the performance of untargeted attacks and compared it with several basic attacks and state-of-the-art attacks.


\bibliography{mybibfile}

\end{document}